\documentclass[10pt, a4paper, numbers=noenddot]{article}
\usepackage{lrec}
\usepackage{multibib}
\newcites{languageresource}{Language Resources}
\usepackage{graphicx}
\usepackage{tabularx}
\usepackage{soul}
\usepackage{epstopdf}
\usepackage{longtable}
\usepackage{tabu}
\usepackage{xargs}
\usepackage{lipsum}
\usepackage[latin1]{inputenc}
\usepackage{hyperref}
\usepackage{xstring}
\usepackage{todonotes}
\usepackage{booktabs}
\usepackage{enumitem}
\usepackage{makecell}
\usepackage{multirow}
\usepackage{tikz}
\usepackage{pgfplots}
\pgfplotsset{compat=1.7}
\usepackage{tabularx}

\newcommand{\secref}[1]{{\getrefnumber{#1}}}
\title{MPST: A Corpus of Movie Plot Synopses with Tags}

\name{Sudipta Kar, Suraj Maharjan, A. Pastor L\'opez-Monroy and Thamar Solorio}

\address{Department of Computer Science\\University of Houston\\Houston, TX 77204-3010\\
         \{skar3, smaharjan2, alopezmonroy, tsolorio\}@uh.edu\\}
\abstract{
Social tagging of movies reveals a wide range of heterogeneous information about movies, like the genre, plot structure, soundtracks, metadata, visual and emotional experiences. Such information can be valuable in building automatic systems to create tags for movies. Automatic tagging systems can help recommendation engines to improve the retrieval of similar movies as well as help viewers to know what to expect from a movie in advance. In this paper, we set out to the task of collecting a corpus of movie plot synopses and tags. We describe a methodology that enabled us to build a fine-grained set of around 70 tags exposing heterogeneous characteristics of movie plots and the multi-label associations of these tags with some 14K movie plot synopses. 
We investigate how these tags correlate with movies and the flow of emotions throughout different types of movies. Finally, we use this corpus to explore the feasibility of inferring tags from plot synopses. We expect the corpus will be useful in other tasks  where analysis of narratives is relevant. 
\newline 
}

\begin{document}
\maketitleabstract
%
%
\section{Introduction}
Folksonomy \cite{vander2005folksonomy}, also known as collaborative tagging or social tagging, is a popular way to gather community feedback about online items in the form of tags.  User-generated tags in  recommendation systems like IMDb\footnote{\url{http://www.imdb.com}} and MovieLens\footnote{\url{https://www.movielens.org}} provide different types of summarized attributes of movies. These tags are effective search keywords, are also useful for discovering social interests, and improving recommendation performance \cite{Lambiotte2006,szomszor2007folksonomies,li2008tag,borne2013collaborative}. In this regard, an interesting research question is: \textit{Can we learn to predict tags for a movie from its written plot synopsis?} This question enables an enormous potential to 
understand the properties of plot synopses that correlate with the tags. For instance, a movie can be tagged with \textit{fantasy, murder,} and \textit{insanity}, that represent different summarized attributes of the movie. The inference of multiple tags by analyzing the written plot synopsis of movies can benefit the recommendation engines. In addition, the consumers would have a useful set of tags representing the plot of a movie. Notwithstanding the usefulness of tags, its proper use in computational methods is challenging as the tag spaces are noisy and redundant \cite{KTV08}. Noise and redundancy issues arise because of differences in user perspectives and use of semantically similar tags. For example, the \textit{Movielens 20M dataset} \citelanguageresource{movielens}, which provides tag assignments between $\approx$27K movies and $\approx$1,100 unique tags also suffers from these problems. Thus, a fine-grained tagset and their assignment to movie plots can help to overcome these obstacles.\\

In this work, (i) we present the MPST corpus that contains plot synopses of 14,828 movies and their associations with a set of 71 fine-grained tags; where each movie is tagged with one or more tags. (ii) We discuss the expected properties of this tagset and present the methodology we followed to create such tagset from multiple noisy tag spaces (Section \secref{sec_data_collect}). We also present the process of mapping these tags to a set of movies and collecting the plot synopses for these movies. (iii) We analyze the correlations between the tags and track the flow of emotions throughout the plot synopses to investigate if the associations between tags and movies fit with what we expect in the real world (Section \secref{sec_data_analysis}). We also try to estimate the possible difficulty level of a multi-label classification approach to predict tags from the plot synopses. (iv) Finally, we create a benchmark system to predict tags using a set of traditional linguistic features extracted from plot synopses. To the best of our knowledge, this is the first corpus that provides multi-label associations between written plot synopses of movies and a fine-grained tagset. The corpus is freely available to download\footnote{\url{http://ritual.uh.edu/mpst-2018}}.

\begin{table}[t]
\setlength\belowcaptionskip{-10pt}
\begin{tabular}{|c|p{0.37\textwidth}|}
\hline
\raisebox{-0.5\height}{\includegraphics[width=0.07\textwidth]{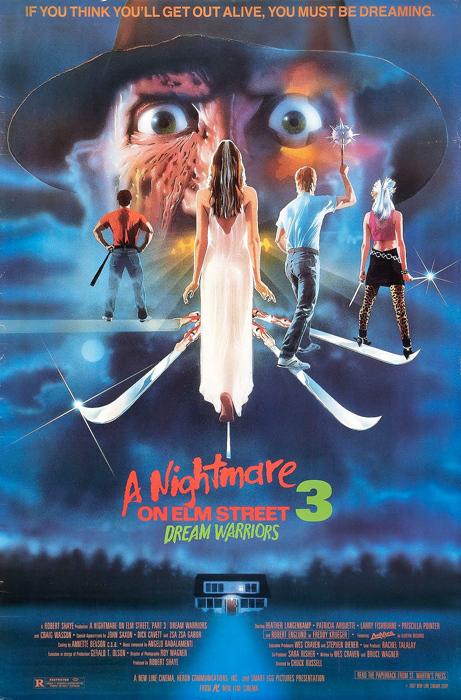}} & 
\makecell{\textbf{A Nightmare on Elm Street 3: Dream} \\ \textbf{Warriors} \\ \textbf{Tags}: \textit{fantasy, murder, cult, violence, horror}, \\ \textit{insanity} }\\
\hline
\raisebox{-0.5\height}{\includegraphics[width=0.07\textwidth]{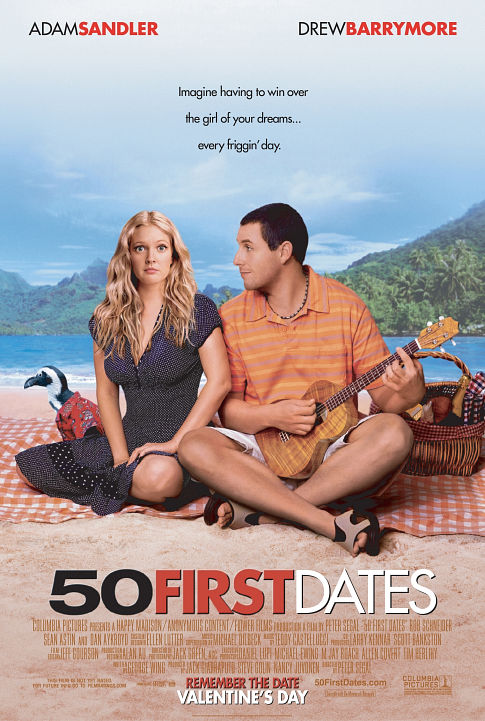}} & 
\makecell{\textbf{50 First Dates} \\ \textbf{Tags}: \textit{comedy, prank, entertaining, romantic}, \\ \textit{flashback}}\\
\hline
\end{tabular}
\caption{\small Examples of tag assignments to movies from the corpus.}
\label{movie_tag_example}
\end{table}
%

%
\section{Creating the Movie Plot Synopses with Tags (MPST) Corpus}\label{sec_data_collect}
There are several datasets that provide plots or scripts of movies. Since their utilization in this work was difficult, we created a fine-grained tagset first and collected the synopses by ourselves.
For example, MM-IMDb \cite{gmu_john_17} provides plot summaries, posters, and metadata of $\approx$25K movies collected from IMDb. But these plot summaries are very short to capture different attributes of movies (average words per summary is 92.5 versus 986.47 in MPST). 
Another example is ScriptBase \cite{scriptbase_gorinsky_2015}, which provides scripts of 1,276 movies collected from IMSDb\footnote{\url{http://www.imsdb.com}}. But plot synopses are more readily available than the scripts and that helped us to create a bigger dataset.
Finally, CMU Movie summary corpus \cite{bamman2014learning} contains $\approx$42K plot synopses of movies collected from Wikipedia. Due to the absence of IMDb id for these movies, we could not retrieve the tag association information for the movies in that corpus.

\begin{figure}
	\centering
	\includegraphics[scale=0.485]{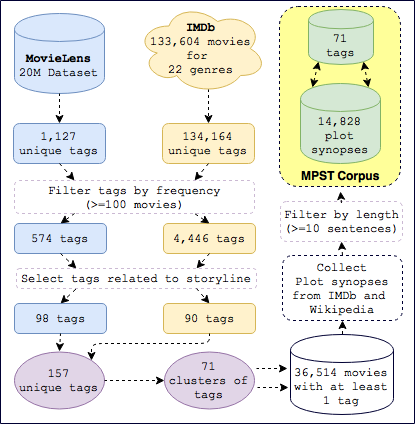}
	\caption{\small{Overview of the data collection process.}}
 	\label{fig:data_collection_chart}
\end{figure}

We created the corpus using \textit{MovieLens  20M dataset}, \textit{Internet Movie Data Base (IMDb)}, and \textit{Wikipedia}. To create a good corpus, we first defined some expected properties of the corpus (Section~\ref{subsec:corpus_property}). Then we created a fine-grained set of tags that satisfies the expected properties (Section \secref{tagset_creation}). We created mappings between the tags and a set of movies and collected the plot synopses for those movies. 
Figure \ref{fig:data_collection_chart} shows an overview of the data collection process that we will discuss in this section.

\subsection{Corpus Requirements} \label{subsec:corpus_property}
We set the following expected properties for the corpus to make it ideal for future works:

\begin{itemize}[leftmargin=0.2cm,labelindent=0cm,noitemsep]
\item \textit{Tags should express plot-related attributes that are easy to understand by people.}\\
The goal is to predict tags from the written movie plots. Therefore relevant tags are those that capture properties of movie plots (e.g. structure of the plot, genre, emotional experience, storytelling style), and not attributes of the movie foreign to the plot, such as metadata.
\item \textit{The tagset should not be redundant.}\\
Because we are interested in designing methods to automatically assign tags, having multiple tags that represent the same property is not desirable. For example, tags like \textit{cult}, \textit{cult film}, \textit{cult movie} are closely related and should all be mapped to a single tag.
\item \textit{Tags should be well represented.}\\
For each tag, there should be a sufficient number of plot synopses, so that the process of characterizing a tag does not become difficult for a machine learning system due to data sparseness.
\item \textit{Plot synopses should be free of noise and adequate in content.}\\
Plot synopses should be free of noise like IMDb notifications and HTML tags. Each synopsis should have at least 10 sentences, as understanding stories from very short texts would be difficult for any learning system.
\end{itemize}

\subsection{Towards a Fine-Grained Set of Tags} \label{tagset_creation}
\begin{figure}
\includegraphics[scale=0.3]{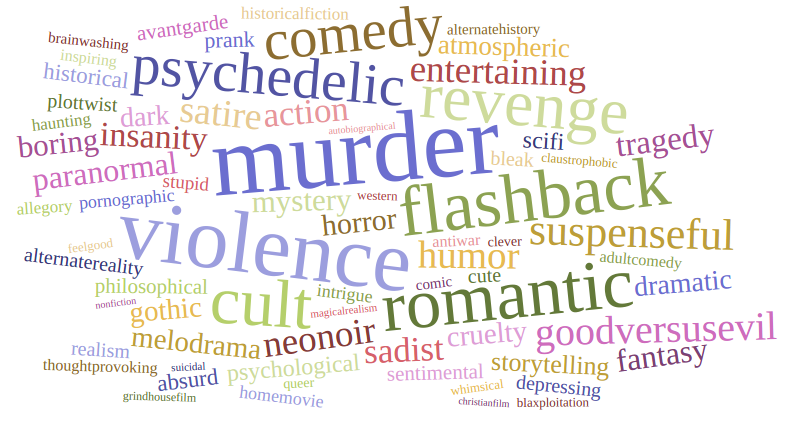}
\caption{\small Tag cloud created by the tags from the dataset. Size of the tags depends on their frequency in the dataset.}
\label{fig:tag_cloud}
\end{figure}

As shown in Figure \ref{fig:data_collection_chart}, we collected a large number of tags from \textit{MovieLens 20M dataset} and \textit{IMDb}. To extract the tags commonly used by the users, we only kept the tags that were assigned to at least 100 movies. We manually examined these tags to shortlist the tags that could be relevant to movie plots. We discarded the tags that do not conform to our requirements. At the next step, we manually examined the tags in this shortlist to group semantically similar tags together. We got 71 clusters of tags by this process and set a generalized tag label to represent the tags of each cluster. For example, \textit{suspenseful}, \textit{suspense}, and \textit{tense} were grouped into a cluster labeled \textit{suspenseful}. Through this step, we overcame the redundancy issues in the tagset and created a more generalized version of the common tags related to the plot synopses. The tagset is shown as a word cloud in Figure \ref{fig:tag_cloud}.\\

We created the mapping between the movies and the 71 clusters using the tag assignment information we collected from \textit{MovieLens 20M dataset} and \textit{IMDb}. If a movie was tagged with one or more tags from any cluster, we assigned the respective cluster label to that movie. We used the IMDb IDs to crawl the plot synopses of the movies from IMDb. We collected synopses from Wikipedia for the movies without plot synopses in IMDb or if the synopses in Wikipedia were longer than the synopses in IMDb. These steps resulted in the MPST corpus that contains 14,828 movie plot synopses where each movie has one or more tags.\\


\section{Data Statistics} \label{sec_data_analysis}
\begin{figure*}
	\centering
	\includegraphics[width=\textwidth]{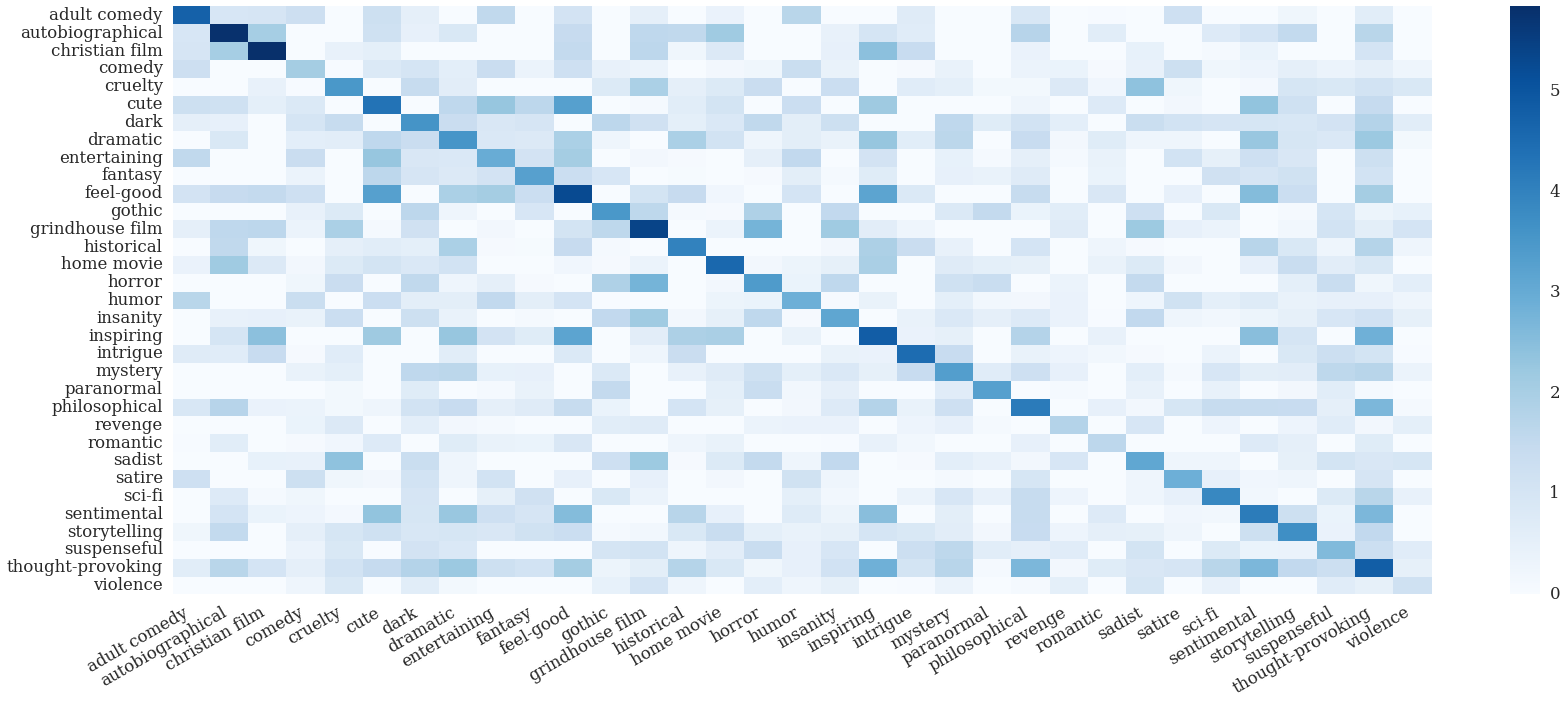}
    \caption{\small{Heatmap of Positive Pointwise Mutual Information (PPMI) between the tags. Dark blue squares represent high PPMI, and white squares represent low PPMI.}}
    \label{fig:tag_tag_pmi_heatmap}
\end{figure*}

\begin{table}[t]
\centering
\scalebox{0.9}{
\begin{tabular}{|l|r|}
\hline
Total plot synopses                      & 14,828 \\
Total tags                               & 71     \\ \hline
Average tags per movie                   & 2.98   \\ 
Median value of tags per movie       	 	 & 2     \\ 
STD of tags for a movie       	 	 & 2.60     \\ 
Lowest number of tags for a movie         & 1      \\ 
Highest number of tags for a movie       & 25     \\ \hline
Average sentences per synopsis    		 & 43.59  \\ 
Median value of sentences per synopsis    	 & 32  \\ 
STD of sentences per synopsis    	 	 & 47.5  \\ 
Highest number of sentences in a synopsis & 1,434   \\ 
Lowest number of sentences in a synopsis  & 10     \\ \hline
Average words per synopsis        & 986.47 \\ 
Median value of words per synopsis        & 728 \\ 
STD of words per synopsis        & 966.16 \\ 
Highest number of words in a synopsis     & 13,576  \\ 
Lowest number of words in a synopsis      & 72     \\ \hline 
\end{tabular}
}
\caption{\small Brief statistics of the MPST corpus.}
\label{tab_data_stat_table}
\end{table}

 Table \ref{tab_data_stat_table} shows that the distribution of the number of tags assigned to movies, number of sentences, and number of words per movie are skewed. Most of the synopses are small in terms of the number of sentences, although the corpus contains some really large synopses with more than 1K sentences. Around half of the synopses have less than 33 sentences. A similar pattern is noticeable for the average number of tags assigned to the movies. Some movies have a large number of tags, but most of the movies are tagged with one or two tags only. \textit{Murder, violence, flashback,} and \textit{romantic} are the most frequent four tags in the corpus that are assigned to 5,732; 4,426; 2,937 and 2,906 movies respectively. Least frequent tags like \textit{non-fiction, christian film, autobiographical, and suicidal} are assigned to less than 55 movies each.

\subsection{Multi-label Statistics}
\textit{Label cardinality} (LC) and \textit{label density} (LD) are two statistics that can influence the performance of multi-label learning methods ~\cite{tsoumakas2006multi,Tsoumakas2010multi2}. Label cardinality is the average number of labels per example in the dataset as defined by Equation~\ref{eq:lc}. 
\begin{equation}
LC(D) = \frac{1}{\left |D  \right |} \sum_{i=1}^{|D|}|Y_i|
\label{eq:lc}
\end{equation}
Here, \(|D|\) is the number of examples in dataset D and \(Y_i\) is number of labels for the \(i^{th}\) example.
Label density is the average number of labels per example in the dataset divided by the total number of labels, as defined by Equation~\ref{eq:ld}.

\begin{equation}
LD(D) = \frac{1}{\left |D  \right |} \sum_{i=1}^{|D|} \frac {|Y_i|}{|L|}
\label{eq:ld}
\end{equation}

Here, \(|L|\) is the total number of labels in the dataset. \newcite{bernardini2014cardinality} analyzed the effects of cardinality and density on multiple datasets. They showed that, for two datasets with similar cardinalities, learning is harder for the one with lower density. And if the density is similar, learning is harder for the one with higher cardinality. For example, learning performance was better for the \textit{Genbase} dataset (LC: 1.252, LD: 0.046) as compared to the \textit{Medical} dataset (LC: 1.245, LD: 0.028), where they had similar cardinalities but the \textit{Medical} dataset was less dense. On the other hand, performance was better for the \textit{Emotions} dataset (LC: 1.869, LD: 0.311) as compared to the \textit{Yeast} dataset (LC: 4.237, LD: 0.303), where they had similar density but cardinality of the \textit{Yeast} dataset was higher. The label cardinality and label density of our  dataset are \textit{2.98} and \textit{0.042}, respectively. Based on the mentioned experiments, we suspect that a traditional multi-label classification approach for this dataset will be a challenge that opens the scope for exploring more scalable approaches.

\begin{figure*}
\includegraphics[width=\textwidth]{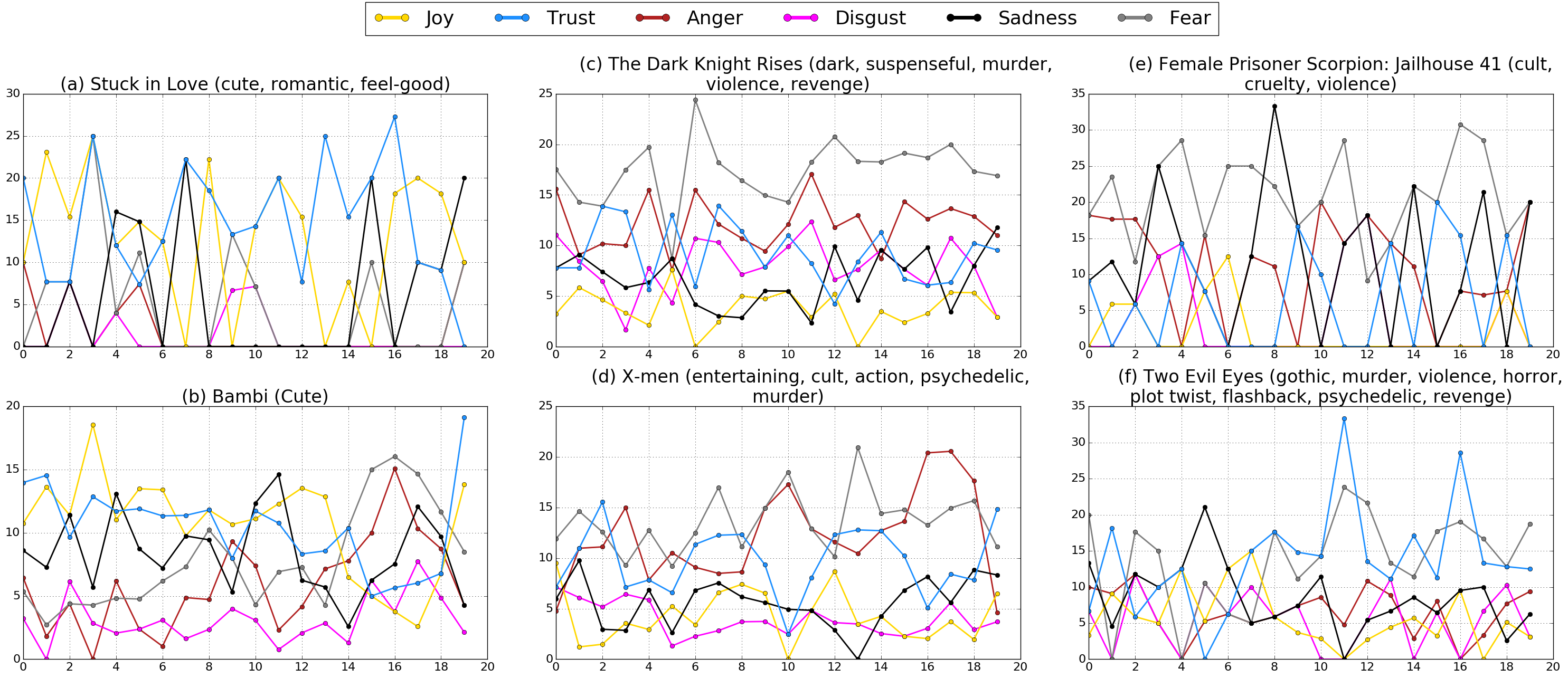}
    \caption{\small Tracking flow of emotions in the synopses of six movies. Each synopsis was divided into equally sized 20 segments based on the words and percentage of the emotions for each segment were calculated using NRC emotion lexicons. The y axis represents the percentage of emotions in each segment; whereas, the x axis represents the segments.}
    \label{fig:emotion_flow}
\end{figure*}

\subsection{Correlation between Tags}

To find out significant correlations in the tagset, we compute the Positive Pointwise Mutual Information (PPMI) between the tags, which is a modification over the standard PMI \cite{pmiChurch:1990:WAN:89086.89095,ppmiDagan:1993:CWS:981574.981596,ppmiNiwa:1994:CVC:991886.991938}. PPMI between two tags \(t1\) and \(t2\) is computed by the following equation:
\begin{equation}
PPMI(t1;t2) \equiv max(log_2 \frac{P(t1,t2)}{P(t1)P(t2)}, 0)
\end{equation}

where, $P(t1,t2)$ is the probability of tags $t1$ and $t2$ occurring together and $P(t1)$ and $P(t2)$ are the probabilities of tag $t1$ and $t2$, respectively.  Figure~\ref{fig:tag_tag_pmi_heatmap} shows the heatmap correlation of PPMI values between a subset of tags. The figure shows interesting relations between the tags and supports our understanding of the real world scenario.\\

High PPMI scores show that \textit{cute, entertaining, dramatic,} and \textit{sentimental} movies can evoke \textit{feel-good} mood, whereas lower PPMI scores between \textit{feel-good} and \textit{sadist, cruelty, insanity,} and \textit{violence} suggest that these movies usually create a different type of impression on people. Also note that, these movies have stronger relations with \textit{horror, cruelty,} and \textit{darkness} which make them difficult to create the \textit{feel-good} experience. It suggest that people tend to get \textit{inspiration} from \textit{dramatic, thought-provoking, historical,} and \textit{home movies}. \textit{Christian films} and \textit{science fictions} are also good sources of \textit{inspiration}. \textit{Grind-house, Christian,} and \textit{non-fiction} films do not usually have \textit{romantic} elements. \textit{Romantic} movies are usually \textit{cute} and \textit{sentimental}. \textit{Autobiographical} movies usually have \textit{storytelling} style and they are \textit{thought-provoking} and \textit{philosophical}. These relations, in fact, show that the movie tags within our corpus seem to portray a reasonable view of movie types based on our understanding of possible impressions from different types of movies.

\subsection{Emotion Flow in the Synopses}
NRC Emotion Lexicons \citelanguageresource{emolex2} have been shown effective to capture the flow of emotions in narrative stories \cite{emotrack}. It is a list of 14,182 words\footnote{Version 0.92} and their binary associations with eight types of elementary emotions from the Hourglass of Emotions model \cite{hourglass_emotion} (\textit{anger, anticipation, joy, trust, disgust, sadness, surprise,} and \textit{fear}) with polarity.\\
In Figure \ref{fig:emotion_flow}, we try to inspect how the flows of emotions look like in different types of plots.
The reason behind this investigation is to get a shallow idea about the potential feasibility of the collected plot synopses to predict tags.
As general users have written the collected plot synopses and created the tags for movies on the web, there is always a possibility to have noise in the data.
For example, in a real world scenario we will expect that horror movies will contain fear and sadness.
On the other hand, comedy or funny movies will be filled with happiness.

In the figure we can observe that, emotions like \textit{joy} and \textit{trust} are dominant over \textit{disgust} and \textit{anger} in \textit{cute, feel-good, and romantic} movie's plots (a, b). We can observe sudden spikes in \textit{sadness} in segment 4. The animated movie \textit{Bambi (1942)} shows an interesting flow of different types of emotions. The dominance of \textit{joy} and \textit{trust} suddenly gets low at segment 14 and gets high again at segment 18, where \textit{fear, sadness,} and \textit{anger} get high at segment 14. It is quite self-explanatory that the plot are mixtures of positive and negative emotions where the lead characters go through difficult situations, fight enemies and face a happy ending (spike in \textit{joy} and \textit{trust} at the end) after climax scenes where enemies get defeated. The final segments of (b) indicate happy endings, but the rise of \textit{sadness} and \textit{fear} in (a) indicates that \textit{Stuck in Love (2012)} does not have a happy ending.

We observe the opposite scenarios in cases of \textit{violent, dark, gothic,} and \textit{suspenseful} movies (c, d, e, and f) where \textit{fear, anger,} and \textit{sadness} dominate over \textit{joy} and \textit{trust}. The dominance of \textit{anger} and \textit{fear} is a good indicator of a movie having \textit{action, violence,} and \textit{suspense}. \textit{Female Prisoner Scorpion: Jailhouse 41 (1972)} (e), has dominance of \textit{fear, sadness,} and \textit{anger} throughout the whole movie, and it is easy to guess that this movie has \textit{violence} and \textit{cruelty} portrayed through the lead characters. The flow of \textit{joy, trust, sadness,} and \textit{fear} alters at the middle of the movie \textit{Two Evil Eyes (1990)} (f). Maybe it is the reason why people tagged it with \textit{plot twist}. These observations give evidence of the connection between the flow of emotion in the plot synopses and the experience people can have from the movies, and they also match with what we  expected.\\

\section{A Machine Learning Approach for Predicting Tags using Plot Synopses}
In this section, we will discuss about some preliminary experiments we conduct with the corpus for predicting tags for movies.
We approach the task of predicting tags for movies as a multi-label classification problem and use various traditional linguistic features.

\subsection{Hand-crafted Features}\label{sec:handcrafted_features}
\noindent {\bf Lexical}: We extract word $n$-grams ($n$=1,2,3), character $n$-grams ($n$=3,4) and two skip $n$-grams ($n$=2,3) from the plot synopses as they are strong lexical representations. We use term frequency-inverse document frequency (TF-IDF) as the weighting scheme.

\noindent {\bf Sentiments and Emotions}:
Sentiments are inherent part of stories and one of the key elements that determine the possible experiences found from a story. For example, \textit{depressive} stories are expected to be full of \textit{sadness, anger, disgust} and negativity, whereas a funny movie is possibly full of \textit{joy} and \textit{surprise}. In this work, we employ two approaches to capture sentiment related features.

\begin{itemize}
\item \noindent {\bf Bag of Concepts}: As concept-level information have showed effectiveness in sentiment analysis \cite{concept_sentiment_cambria}, we extract around 10K unique concepts from the plot synopses using the Sentic Concept parser\footnote{\url{https://github.com/SenticNet/concept-parser}}. It breaks sentences into verb and noun clauses and extracts concepts from them using Parts of Speech (POS) based bigram rules \cite{rajagopal2013graph}.

\item \noindent {\bf Affective Dimensions Scores}: The hourglass of emotions model \cite{hourglass_emotion} categorized human emotions into four affective dimensions (\textit{attention, sensitivity, aptitude and pleasantness}) starting from the study on human emotions by \newcite{plutchik2001nature}. Each of these affective dimensions is represented by six different activation levels called `sentic levels'. These make up to 24 distinct labels called `elementary emotions' that represent the total emotional state of the human mind. SenticNet 4.0 \cite{senticnet4} knowledge base consists of 50,000 common-sense concepts with their semantics, polarity value and scores for the basic four affective dimensions. We used this knowledge base to compute average polarity, attention, sensitivity, aptitude, and pleasantness for the synopses.


\end{itemize}
We divide the plot synopses into three equal chunks based on words and extracted these two sentiment features for each chunk. We will discuss more about chunk-based sentiment representation later.

\noindent {\bf{Semantic Frames}}: Semantic role labeling is a useful technique to assign abstract roles to the arguments of predicates or verbs of sentences. 
We use SEMAFOR\footnote{\url{http://www.cs.cmu.edu/~ark/SEMAFOR}} frame-semantic parser to parse the frame-semantic structure using the FrameNet \cite{baker1998berkeley} frames. 
For each synopsis, we use the bag of frames representation weighted by normalized frequency as feature.

\noindent {\bf Word Embeddings}: Word embeddings have been shown effectiveness in text classification problems by capturing semantic information. Hence, in order to capture the semantic representation of the plots, we average the word vectors of every word in the plot. We use the publicly available FastText pre-trained word embeddings\footnote{\url{https://github.com/facebookresearch/fastText/blob/master/pretrained-vectors.md}}.

\noindent {\bf{Agent Verbs and Patient Verbs}}: Actions done and received by the characters can help to identify attributes of plots. For example, if the characters of a movie \textit{kill, take revenge, shoot, smuggle, chase}; we can expect violence, murder, action from that story. We use the agent and patient verbs found in synopses to capture the actions. In this regard, we use Stanford CoreNLP library to parse the dependencies of the synopses. Then we extract the agent verbs (using \textit{nsubj or agent} dependencies) and the patient verbs (using \textit{dobj, nsubjpass, iobj} dependencies) as described in \newcite{bamman2014learning}. 
We group these verbs into 500 clusters using the pre-trained word embeddings with the K-means clustering algorithm to reduce noise.
We use the distribution of these clusters of the agent verbs and patient verbs over the synopses.
We experimented with different values of K (K=100, 500, 1000, 1500), and 500 clusters helped to achieve better results. 

\subsection{Experimental Setup}
Section {\secref{sec_data_analysis}} shows that the distribution of the number tags assigned to per movies is skewed. The average number of tags per movie is approximately three. We  thus begin by experimenting with predicting a fixed number of three tags for each movie. Moreover, to get more detailed idea about movies, we create another set of five tags by predicting two additional tags.

We use random stratified split to divide the data into 80:20 train to test ratio\footnote{Train-test partition information is available with the dataset.}. We use the One-versus-Rest approach to predict multiple tags for an instance. We experiment with logistic regression as the base classifier. 
We run five-fold cross-validation on the training data to evaluate different features and combinations. We tune the regularization parameter ($C$) using  grid search technique over the best feature combination that includes all of the extracted features.
We use the best parameter value (\textit{C=0.1}) for training a model with all the training data and used that model for predicting tags for the test data. 

\noindent{\textbf{Majority and Random Baseline:}} We define majority and random baselines to compare the performance of our proposed model in the task of predicting tags for movies. The majority baseline method assigns the most frequent three or five tags to all the movies. We chose three tags per movie as this is the average number of tags per movie in the dataset. Similarly, the random baseline assigns at random three or five tags to each movie. 

\noindent{\textbf{Evaluation Metrics:}} \newcite{multilabel_metrics_wu} illustrate the complications in evaluating multi-label classifiers by an example of determining  the significance of mistakes for the following cases: \textit{one instance with three incorrect labels} vs. \textit{three instances each with one incorrect label}.
It is complicated to tell which of these mistakes is more serious.
Due to such complications, several evaluation methodologies have been proposed for this type of tasks \cite{tsoumakas2006multi,multilabel_metrics_wu}. For example, \textit{hamming loss, average precision, ranking loss, one-error, coverage, }\cite{schapire2000boostexter,multilabel_calibrated_frnkranz_2008}, \textit{micro} and \textit{macro} averaged versions of \textit{F1} and \textit{AUC} score \cite{Tsoumakas2010multi2,tsoumakas2011random,multi_auc_diag_lipton}.\\
%
Another complication arises when the label distribution is sparse in a dataset. Less frequent tags could be under-represented by models, but an ideal model should be able to discriminate among all the possible labels. Such an issue is very common in problems like image annotation, and existing works use \textit{mean per label recall} and \textit{labels with recall}$>$0 to measure the effectiveness of models in learning individual labels \cite{img_annot5_Lavrenko,img_annot3_Feng,img_annot1_Carneiro,wang09-sparse}. Here, we use two similar metrics: \textit{tag recall} (TR) and \textit{tags learned} (TL), along with traditional micro-F1 metric. 
Tag recall computes the average recall per tag and defined by the following equation.
\begin{equation}
TR = \frac{\sum_{i=1}^{|T|}|R_i|}{|T|}
\label{eq:tr}
\end{equation}
Here, $|T|$ is the size of tagset in the corpus, and $R_i$ is recall of $i^{th}$ tag.
Tags learned (TL) computes how many unique tags are being predicted by the system for the test data.
These evaluation metrics will help us to investigate how well and how many distinct tags are being learned by the models. We evaluate the models using these three metrics in two settings. One is selecting the top three tags and another is selecting the top five tags.

\subsection{Results and Analysis}
\pdfoutput=1
\begin{table}[]
\centering
\resizebox{\columnwidth}{!}{%
\begin{tabular}{@{}|l|llr|llr|@{}}
 \hline
 & \multicolumn{3}{c|}{\textbf{Top 3}} & \multicolumn{3}{c|}{\textbf{Top 5}} \\  \hline
 & \textbf{F1} & \textbf{TR} & \textbf{TL} & \textbf{F1} & \textbf{TR} & \textbf{TL} \\  \hline
Baseline: Most Frequent & 29.7 & 4.225 & 3 & 31.5 & 7.042 & 5 \\
Baseline: Random & 4.20 & 4.328 & \textbf{71} & 5.40 & 7.281 & \textbf{71} \\ \hline
Unigram (U)                  & 37.6   & 7.883    & 22.6  & 37.1   & 11.945   & 27.4  \\
Bigram (B)                   & 36.5   & 7.216    & 19.6  & 36.1   & 10.808   & 24.8  \\
Trigram (T)                  & 31.3   & 5.204    & 15.4  & 32.4   & 8.461    & 21    \\
Char 3-gram (C3)             & 37.0   & 7.419    & 22    & 36.6   & 11.264   & 27.4  \\
Char 4-gram (C4)             & 37.7   & 7.799    & 22.6  & 37.0   & 11.582   & 27.2  \\
2 skip 2 gram (2S2)          & 34.2   & 6.289    & 19.4  & 34.5   & 9.875    & 25.2  \\
2 skip 3 gram (2S3)          & 30.8   & 4.951    & 12.8  & 32.1   & 8.109    & 18.2  \\
Bag of Concepts (BoC)         & 35.7   & 7.984    & 29    & 35.9   & 12.473   & 34.8  \\
Concepts Scores (CS)        & 31.1   & 4.662    & 7.8   & 32.4   & 7.512    & 8.2   \\
Word Embeddings              & 36.8   & 6.744    & 13.2  & 36.1   & 10.074   & 17.8  \\
Semantic Frame               & 33.4   & 5.551    & 13.4  & 33.9   & 8.394    & 15.2  \\ 
Agent Verbs      & 32.9   & 5.050    & 7.2   & 33.2   & 7.714    & 8     \\
Patient Verbs    & 33.1   & 5.134    & 7.4   & 33.5   & 7.843    & 8     \\ \hline
U+B+T                        & 37.2   & 8.732    & 30    & 36.8   & 13.576   & 36.8  \\
C3+C4                        & \textbf{37.8}   & 8.662    & 28.8  & \textbf{37.4}   & 13.395   & 33.6  \\
U+B+T+C3+C4                  & 37.1   & 9.991    & 36.8  & 36.8   & 15.871   & 45.8  \\
Al lexical         & 36.7   & 10.046   & 37.6  & 36.5   & 15.838   & 46.4  \\
BoC + CS    & 35.7   & 8.165    & 29.4  & 36.0   & 12.754   & 35.4  \\
\hline



\hline
\textbf{All features }& 36.9 & \textbf{10.364} & 39.6 & 36.8 & \textbf{16.271} & 47.8 \\  \hline 
\end{tabular}%
}
\caption{Performance of the hand-crafted features using 5-fold cross-validation on the training data. We use three metrics (\textit{F1: micro averaged F1, TR: tag recall, and TL: tags learned}) to evaluate the features.}\label{tab:tab_results_1}
\end{table}
\pdfoutput=1
\begin{table}[t]
\centering
\resizebox{\columnwidth}{!}{%
\begin{tabular}{@{}|l|llr|llr|@{}}
 \hline
 & \multicolumn{3}{c|}{\textbf{Top 3}} & \multicolumn{3}{c|}{\textbf{Top 5}} \\  \hline
 & \textbf{F1} & \textbf{TR} & \textbf{TL} & \textbf{F1} & \textbf{TR} & \textbf{TL} \\  \hline
Baseline: Most Frequent & 29.7 & 4.23 & 3 & 28.4 & 14.08 & 5 \\
Baseline: Random & 4.20 & 4.21 & \textbf{7}1 & 6.36 & 15.04 & \textbf{71} \\ \hline

\textbf{System}& \textbf{37.3} & \textbf{10.52} & 47 & \textbf{37.3} & \textbf{16.77} & 52 \\  \hline 
\end{tabular}%
}
\caption{Results achieved on the test data using the best feature combination (all features) with tuned regularization parameter \textit{C}.}\label{tab:tab_test_results}
\end{table}
Table \ref{tab:tab_results_1} shows the performance of the hand-crafted features for predicting tags for movies. All the features beat the baselines in terms of micro-F1 and tag recall (TR). But another significant criterion to evaluate the performances is the number of unique tags predicted by the models, which is measured by the tags learned (TL) metric. We prefer such a model that is capable of creating diverse tagsets by capturing varieties of attributes of movies with reasonable accuracy. 
For instance, the random baseline used all of the tags in the dataset to assign to the movies but its accuracy is very poor. 
On the other hand, the majority baseline has better accuracy but it does not have diversity in the tagset.
We can see that most of the individual features achieved almost similar micro-F1 scores, but they demonstrate difference in effectiveness to create diversity in predicted tags. Feature combinations seem to improve in TR and TL, but micro-F1 scores are almost similar to the individual features.\\
The lexical features show better performance compared to other features.
Bag of concepts (BoC) shows similarity in performance.
Combination of all lexical features demonstrates effectiveness in capturing a wide range of attributes of movies from the synopses, which is reflected by the better TR and TL scores.\\
We present the results achieved on the test data in Table \ref{tab:tab_test_results}. Although the result is similar to the result we got with all features during cross-validation, number of predicted unique tags is higher in the test set. This result could be used as a baseline system to compare other methods developed in future as it uses several traditional linguistic features combination to predict tags.\\
\noindent{\textbf{Chunk-based Sentiment Representation: }} Narratives have patterns in ups and downs of sentiments \cite{sunday1981autobiographical}. \newcite{emotional_arc_reagan2016} showed that the pattern of changes in sentiments  is significant for consumer experiences that results in success of stories.  
To capture such changes, we experiment with chunk-based sentiments and emotions representation. 
We divide the plot synopses into equally sized $n$ chunks based on the word tokens and extract the sentiment and emotion features for each chunk.
Then we run 5-fold cross validation on the training data to observe the effect of chunk-based sentiments and emotions representation.
We report the results in Table \ref{tab:tab_results_chunk}.
Results show that dividing synopses into multiple chunks and using sentiment and emotion features from each chunk improves the performance of tag prediction.
Although we observe noticeable improvements up to three chunks, TL remains similar where micro-F1 scores start to drop when we use more than three chunks.
We suspect that higher number of chunks create sparseness in the representation of sentiments and emotions that hurts the performance.
So we use sentiments and emotions features using three chunks in further experiments.
As the chunk-based representation shows improvement in results, we plan to work capturing the flow of sentiments throughout the plots more efficiently in future work.
\pdfoutput=1
\begin{table}[t]
\centering
\resizebox{\columnwidth}{!}{%
\begin{tabular}{@{}|l|llr|llr|@{}}
 \hline
\multirow{2}{*}{\textbf{Chunks}} & \multicolumn{3}{c|}{\textbf{Top 3}}           & \multicolumn{3}{c|}{\textbf{Top 5}}           \\ \cline{2-7} 
                                                      & \textbf{F1} & \textbf{TR} & \textbf{TL} & \textbf{F1} & \textbf{TR} & \textbf{TL} \\ \hline
1     & 35.2   & 6.550    & 18.2  & 35.1   & 9.928    & 23.4  \\
2     & 35.0   & 7.031    & 23.0   & 35.2   & 10.68    & 26.8  \\
3     & \textbf{35.7}   & 8.165    & 29.4  & \textbf{36.0}   & \textbf{12.754}   & 35.4  \\
4     & 35.1   & 8.153    & 30.6  & 35.4   & 12.723   & \textbf{36.8}  \\
5     & 34.8   & \textbf{8.185}    & 30.4  & 35.1   & 12.553   & \textbf{36.8} \\
6     & 34.3   & 7.976    & \textbf{31.2}  & 34.9   & 12.725   & 36.0    \\ 
\hline
\end{tabular}%
}
\caption{Experimental results obtained by 5-fold cross-validation using chunk-based sentiment representations. Chunk-based sentiment features were combined with the other features described in Section \ref{sec:handcrafted_features}}\label{tab:tab_results_chunk}
\end{table}\\
%
%
\section{Conclusion}
We have presented a new corpus of $\approx$70 fine-grained tags and their associations with $\approx$14K plot synopses of movies. In order to create the tagset, we tackled the challenge of extracting tags related to movie plots from noisy and redundant tag spaces created by user communities in MovieLens and IMDb. In this regard, we describe the methodology for creating the fine-grained tagset and mapping the tags to the plot synopses. \\
We present an analysis, where we try to find out the correlations between tags. These correlations seem to portray a reasonable set of movie types based on what we expect from certain types of movies in the real world. We also try to analyze the structure of some plots by tracking the flow of emotions throughout the synopses, where we observed that movies with similar tag groups seem to have similarities in the flow of emotions throughout the plots.\\
Finally, we create a benchmark system to predict tags from the synopses using a set of hand-crafted linguistic features.
This dataset will be helpful to analyze and understand the linguistic characteristics of plot synopses of movies, which will in turn help to model certain types of abstractions as tags. For example, what type of events, word choices, character personas, relationships between characters, and plot structure make a movie \textit{mysterious} or \textit{suspenseful} or \textit{paranormal}? Such investigations can help the research community to better exploit high-level information from narrative texts, and also help to build automatic systems to create tags for movies. The generation of tags from movie plots or narrative texts could also be a significant step towards solving the problem of automatic movie profile generation. Methodologies designed using the MPST corpus could also be used to analyze narrative texts from other domains, such as books and storyline of video games.\\

\section*{Acknowledgements}
We would like to thank the National Science Foundation for partially funding this work under award 1462141. We are also grateful to Prasha Shrestha, Giovanni Molina, Deepthi Mave, and Gustavo Aguilar for reviewing and providing valuable feedback during  the  process of creating tag clusters.

\section{Bibliographical References}
\label{main:ref}

\bibliographystyle{lrec}
\bibliography{xample}

\section{Language Resource References}
\label{lr:ref}
\bibliographystylelanguageresource{lrec}
\bibliographylanguageresource{xample}


\end{document}